\documentclass[runningheads]{llncs}
\usepackage{graphicx}
\usepackage{subcaption}
\usepackage{multirow}
\begin{document}
\title{L3Cube-MahaHate: A Tweet-based Marathi Hate Speech Detection Dataset and BERT models\thanks{Supported by L3Cube Pune}}
\titlerunning{L3Cube-MahaHate: A Marathi Hate Speech Detection Dataset}
% If the paper title is too long for the running head, you can set
% an abbreviated paper title here
%
\author{Abhishek Velankar\inst{1,3} \and
Hrushikesh Patil\inst{1,3} \and
Amol Gore\inst{1,3} \and
Shubham Salunke\inst{1,3} \and
Raviraj Joshi\inst{2,3}
% \orcidID{1111-2222-3333-4444}
}
\authorrunning{A. Velankar et al.}
% First names are abbreviated in the running head.
% If there are more than two authors, 'et al.' is used.
%
\institute{Pune Institute of Computer Technology, Pune, Maharashtra India \and
Indian Institute of Technology Madras, Chennai, Tamilnadu India \and
L3Cube Pune \\
\email{\{velankarabhishek,hrushi2900,amolgore960,shubhamsalunke30012001\}@gmail.com, }\\
% \url{http://www.springer.com/gp/computer-science/lncs} \and
% ABC Institute, Rupert-Karls-University Heidelberg, Heidelberg, Germany\\
\email{ravirajoshi@gmail.com}}
\maketitle              % typeset the header of the contribution
\begin{abstract}
Social media platforms are used by a large number of people prominently to express their thoughts and opinions. However, these platforms have contributed to a substantial amount of hateful and abusive content as well. Therefore, it is important to curb the spread of hate speech on these platforms. In India, Marathi is one of the most popular languages used by a wide audience. In this work, we present L3Cube-MahaHate, the first major Hate Speech Dataset in Marathi. The dataset is curated from Twitter, annotated manually. Our dataset consists of over 25000 distinct tweets labeled into four major classes i.e hate, offensive, profane, and not. We present the approaches used for collecting and annotating the data and the challenges faced during the process. Finally, we present baseline classification results using deep learning models based on CNN, LSTM, and Transformers. We explore mono-lingual and multi-lingual variants of BERT like MahaBERT, IndicBERT, mBERT, and xlm-RoBERTa and show that mono-lingual models perform better than their multi-lingual counterparts. The MahaBERT model provides the best results on L3Cube-MahaHate Corpus.
The data and models are available at {\footnotesize https://github.com/l3cube-pune/MarathiNLP}\\

\keywords{Hate Speech Detection \and Marathi Dataset \and Marathi NLP \and Transformers \and MahaBERT}
\end{abstract}

\section{Introduction}

In the past decade, there has been an expeditious rise in the popularity of online social media platforms all over the globe. People have become more open to sharing their opinions without thinking excessively. This often leads to the spread of hate or offensive speech thereby causing violence and cyberbullying. Hate speech is a kind of abusive language directed towards a community that is underprivileged in terms of race, gender, ethnic origin, disability, etc., or can be an insult or threat to an individual \cite{macavaney2019hate,matamoros2021racism}. The users often defy the boundaries of freedom of speech without even realizing it by posting harmful messages and comments \cite{waseem2016hateful}. Therefore it is today's need to neutralize these activities from proliferating further.
\\ \\
In this work, we consider hate speech detection in the Marathi language, a regional language in India, spoken by over 83 million people across the country \cite{joshi2022l3cube}. Despite being one of the popular languages in India, work in the area of hate speech detection in Marathi is extremely limited \cite{mandl2021overview,velankar2021hate,glazkova2021fine,bhatia2021one} as compared to other languages \cite{del2017hate,romim2021hate,corazza2020multilingual,schmidt2019survey}. Even general text classification in Marathi has received limited attention \cite{kulkarni2022experimental,kulkarni2021l3cubemahasent}. In this paper, we present, L3Cube-MahaHate Corpus, the largest publicly available hate speech dataset in Marathi. The dataset is collected from Twitter, tagged with four fine-grained labels which are defined as follows -
\\ \\
\textit{\textbf{Hate (HATE):}} A Twitter post abusing a specific group of people or community based on their religion, race, ethnic origin, gender, geographical location, etc. stimulating violent behaviors.
\\ \\
\textit{\textbf{Offensive (OFFN):}} A tweet containing harmful language leading to insulting or dehumanizing, at times threatening a particular individual.
\\ \\
\textit{\textbf{Profane (PRFN):}} A tweet including the use of typical swear words or profane, cursing language which is ordinarily insupportable.
\\ \\
\textit{\textbf{Not (NOT):}} A post that does not contain any insulting or abusive content or profane words and appears normal in terms of the language used.
\\ \\
The dataset consists of over 25000 samples tagged manually with the classes explained above. We further provide an extensive study of the data collection approaches, different policies used, and challenges faced during the annotation process as well.  We also provide the statistical analysis of our dataset along with the distribution of train, test, and validation data. Lastly, we perform multiple experiments to evaluate state-of-the-art deep learning models on the dataset and provide the baseline results to the community.

The MahaBERT model fined-tuned on L3Cube-MahaHate is termed as MahaHate-BERT\footnote{https://huggingface.co/l3cube-pune/mahahate-bert}\footnote{https://huggingface.co/l3cube-pune/mahahate-multi-roberta} and is shared publicly on model hub. All the resources are publicly shared on github\footnote{https://github.com/l3cube-pune/MarathiNLP}.

\section{Related Work}

Hate speech detection is considered to be a highly critical problem and a lot of attempts have been made to control it. A significant amount of work can be seen in English text analysis. But recently, efforts have been made towards widening the research in regional languages like Marathi as well.
\\ \\
\cite{gaikwad2021crosslingual} presented the Marathi Offensive Language Dataset (MOLD), with nearly 2,500 annotated tweets labeled as offensive and not offensive. It is considered the first dataset for offensive language identification in Marathi. Also, they evaluated the performance of several traditional machine learning models and deep learning models (e.g. LSTM) trained on MOLD.
\\ \\
\cite{bhardwaj2020hostility} collected over 8200 hostile and non-hostile Hindi text samples from multiple social media platforms like Twitter, Facebook, WhatsApp. Hostile posts were further extended into fake, defamation, hate, and offensive. Total 8192 posts were collected and tested on various machine learning models using mBERT encoding.
\\ \\
A Hindi-English code-mixed corpus was constructed in \cite{bohra2018dataset} using the tweets posted online for the duration of five years. Tweets were scrapped using Twitter python API by selecting certain hashtags and keywords from political events, public protests, riots, etc. After removing noisy samples a dataset of 4575 code-mixed tweets was created. The experiments were performed with SVM and Random Forest algorithms along with character and word N-gram features.
\\ \\
In \cite{kulkarni2021l3cubemahasent} authors presented a dataset containing over ~16000 Marathi tweets, manually tagged in three classes namely positive, negative and neutral. They also provided a policy for tagging sentences by their sentiment. Analysis was performed on CNN, BiLSTM, and BERT models.
\\ \\
\cite{davidson2017automated} collected hate phrases identified by Hatebase.org, then used those phrases to collect English tweets from Twitter using Twitter API. The final set of 25k tweets was annotated by CrowdFlower workers with labels hate, offensive and neither. This dataset was then tested on  Logistic Regression, Naive Bayes, Decision Trees, random forests, and linear SVMs.
\\ \\
In \cite{geetdsa:hal-03244472}, the authors evaluated the effect of filtering the generated data used for Data Augmentation (DA). This demonstrates up to 7.3\% and up to 25\% of relative improvements on macro-averaged F1 on two widely used hate speech corpora.
\\ \\
\cite{8683170} proposed a hypothesis that there exists a relation between fake messages or rumors and sentiments of the texts posted online. The experiments were performed on the standard Twitter fake news dataset and showed good improvement on the same.
\\ \\
\cite{gao2018detecting} provided an annotated corpus of hate speech with the context information. This evaluates by using logistic regression and neural network models for hate speech detection around 3\% and 4\%, and it improves to 7\% by combining these two models together.
\\ \\
\cite{mathur2018did} presented MIMCT to detect offensive(Hate or Abusive) Hinglish tweets from the proposed Hinglish Offensive Tweet dataset. Demonstrated the use of the multi-channel CNN-LSTM model for sentiment analysis.
\\

\section{Dataset Creation}
\subsection{Collection}

We created the Hate Speech dataset using the tweets posted online by different users across the Maharashtra region considering the period of over the last 5 years. There are plenty of different python libraries available such as Twint, GetOldTweets, Snscrape, etc. which can be used to collect Twitter posts. Twitter provides its own API as well. We used the Twint python library for scraping the tweets.
\\ \\
To obtain the hateful tweets, firstly, we created a list of over 150 bad words in Marathi which are predominantly used by online users to spread hostility. Some of these are typical swear words in Marathi and other offensive words. These words were in Marathi Devanagari script as we are not concerned about Roman or code-mixed text in this work. We will be publishing the final list on GitHub. These words were used as a search query to obtain hate, offensive, and profane tweets. The majority of the tweets that we obtained are related to political and social issues. We also made a note of controversial events with their time frame happening in India in the last couple of years which particularly triggered violence on social media. 
% We focused on a few user accounts which regularly post hateful content and a few hashtags as well. 
To avoid bias towards certain words or phrases, we have limited the tweets for a particular search query to a number less than 150. Also, while collecting the tweets, we have not included any reference to the author of the tweet thereby eliminating the bias towards that author.
\\ \\
In our publicly available version of the dataset, we have kept all the hashtags, symbols, emojis, and URLs for anyone to experiment on. However, we have removed all of these while performing the baseline experiments. Furthermore, we will be removing the user mentions from the public dataset to maintain complete user anonymity. 

\subsection{Annotation}

The entire dataset has been labelled manually by the 4 annotators considering four major classes viz. hate, offensive, profane, and not. All the annotators were native Marathi speakers and were fluent in reading and writing in Marathi. The annotation guidelines were set before the tagging exercise. The first 200 sentences were tagged together to further improve the consistency post which sentences were tagged in parallel except for ambiguous sentences. The tweets which were targeted at a single individual thereby criticizing or dehumanizing the individual are tagged as offensive. These tweets were mainly attributed to an individual politician, celebrity, or any random person with the use of singular phrases. The tweets which were targeted at a group of people describing the deficiencies towards race, political opinion, sexual orientation, gender, etc. are tagged as hate. These tweets were majorly concentrated towards political parties or the ruling government. Also, a few samples belong to negative comments on minority groups and gender bias. The tweets which contain swear or profane words are strictly tagged as profane, even if they describe the offensive or hateful category. The tweets that do not satisfy any of the above criteria are simply tagged as not. Congratulatory and thanking tweets are tagged as not as well.
\\ \\
In some cases, the intention of the user behind a tweet cannot be suitably identified. In such cases, the tweets were reviewed again and voting among 4 annotators was used to decide on the labels. Also, we encountered a few tweets where hateful comments were quoted by a news handle. As these posts may indirectly promote violence, we tagged them in the hateful category. To collect the NOT tweets, we selected some Marathi personalities and scraped their tweets, which gave us unbiased data towards any word.
\\ \\
\begin{figure}[hbt!]
    \centering
    \includegraphics[width=7cm]{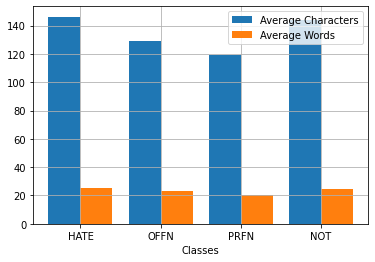}
    \caption{Average characters and words per label}
    \label{fig:avg}
\end{figure}
\begin{table*}[hbt!]
\centering
\begin{tabular}{c c c c c c}
\hline
\textbf{Split}&\textbf{HATE}&\textbf{OFFN}&\textbf{PRFN}&\textbf{NOT}&\textbf{TOTAL} \\
\hline
Train & 5375 & 5375 & 5375 & 5375 & 21500 \\
Test & 500 & 500 & 500 & 500 & 2000 \\
Validation & 375 & 375 & 375 & 375 & 1500 \\
\hline
\end{tabular}
\caption{\label{table:1}
Dataset label distribution
}
\end{table*}
\subsection{Dataset Details}
\begin{figure}[hbt!]
    \centering
    \includegraphics[width=7cm]{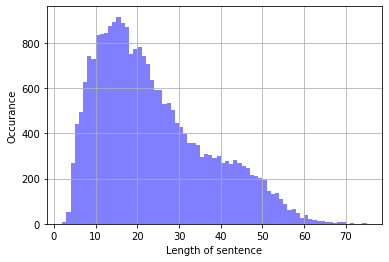}
    \caption{Distribution of the length of a tweet}
    \label{fig:length}
\end{figure}

\begin{figure}[hbt!]
    \centering
    \includegraphics[width=7cm]{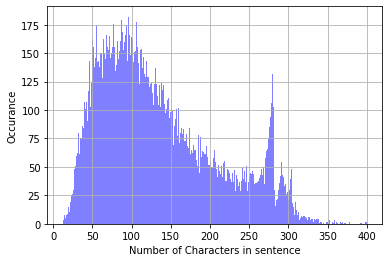}
    \caption{Distribution of the number of characters in a tweet}
    \label{fig:chars}
\end{figure}

Initially, we collected over 40k tweets in Marathi. Among these, we annotated {\raise.17ex\hbox{$\scriptstyle\sim$}}28000 samples. After removing over 3k noisy tweets which particularly included poorly written text i.e. the text with the use of regional words or a large amount of grammatical mistakes, we randomly selected 6250 samples from each of the 4 classes giving the total count of 25000 tweets. Although this uniform distribution of tweets does not represent the true distribution it makes the model building easier and does not require imbalance handling. We analyzed a few statistics on the dataset. The average number of words per tweet in an entire dataset is 21 and the average number of characters is 113. The label-wise distribution is given in Figure \ref{fig:avg}. The length of samples varies in the range of 2 to 93. The distribution of the length of tweets and the number of characters per tweet is given in Figures \ref{fig:length} and \ref{fig:chars} respectively. The dataset can be used for binary classification as well. To match the number of hateful samples viz. Hate, Offensive, Profane all included, we collected over 12500 extra NOT samples apart from that of 4-class corpus giving an equal distribution of 18750 samples in hateful and non-hateful categories. This binary corpus of 37.5k will also be provided along with the original dataset. Table \ref{table:1} shows the 4-class dataset distribution in training, testing and validation samples.
\\
\\

\section{Experiments}

\subsection{Model architectures}

We have used multiple state of the art deep learning architectures \cite{velankar2021hate}, \cite{2021}, \cite{joshi2019deep} to obtain the baseline results on 2-class as well as 4-class classification. Before training the models, we have cleaned the data by removing unwanted symbols, user mentions, hashtags. Following algorithms are used for the evaluation of results:
\\ \\ 
\textbf{CNN}: The CNN model has a 1D convolution layer with a filter of size 300 and a kernel of size 3. It used ReLU activation, followed by max-pooling with pool size 2. the same layers were added again which is followed by a dense layer of size 50 and ReLU activation. Lastly, the layer with softmax activation and 2 nodes was used. A dropout of 0.3 was used after the 1D max-pooling layer. 
\\ \\
\textbf{LSTM}: The LSTM layer with 32 nodes was used. It was followed by a 1D global max-pooling. The dense layer with 16 nodes along with ReLU activation was used, followed by 0.2 dropout. A dense layer with 2 nodes and softmax activation was used as a final layer of the model. 
\\ \\
\textbf{BiLSTM}: Bi-LSTM layer with 300 nodes followed by a 1D global max-pooling layer was used. The dense layer was used with 100 nodes and ReLU activation was used with it. This was followed by a dropout of 0.2. At last, the final layer with 2 nodes with activation softmax was used. 
\\ \\
\textbf{BERT}: BERT is a bi-directional transformer based model \cite{devlin2019bert} pre-trained over large textual data to learn language representations. It can be fine-tuned for specific machine learning tasks. We used the following variations of BERT to obtain baseline results:

\begin{itemize}

\item Multilingual-BERT (mBERT) - trained on and usable with 104 languages with Wikipedia using a masked language modeling (MLM) objective \cite{DBLP:journals/corr/abs-1810-04805}.
\item IndicBERT - a multilingual ALBERT model released by Ai4Bharat, trained on large-scale corpora \cite{kakwani2020indicnlpsuite}, covering 12 major Indian languages: Assamese, Bengali, English, Gujarati, Hindi, Kannada, Malayalam, Marathi, Oriya, Punjabi, Tamil, Telugu.
\item XLM-RoBERTa - a multilingual version of RoBERTa \cite{DBLP:journals/corr/abs-1911-02116}. It is pre-trained on 2.5TB of filtered CommonCrawl data containing 100 languages with the Masked language modeling (MLM) objective and can be used for downstream tasks.
\item MahaBERT - a multilingual BERT model \cite{joshi2022l3cube} fine-tuned on L3Cube-MahaCorpus and other publicly available Marathi monolingual datasets containing a total of 752M tokens.
\\ 
\end{itemize}

\begin{table*}[hbt!]
\centering
\begin{tabular}{c c c c}
\hline
\textbf{Model}&\textbf{Variant}&\textbf{2-Class Accuracy}&\textbf{4-Class Accuracy} \\
\hline
\multirow{3}{*}{CNN} & Random & \textbf{0.880} & 0.703 \\
 & Trainable & 0.866 & 0.710 \\
 & Non-Trainable & 0.870 & \textbf{0.751} \\
\hline
\multirow{3}{*}{LSTM} & Random & 0.857 & 0.681 \\
 & Trainable & 0.860 & 0.691 \\
 & Non-Trainable & \textbf{0.869} & \textbf{0.751} \\
\hline
\multirow{3}{*}{BiLSTM} & Random & 0.858 & 0.699 \\
 & Trainable & 0.860 & 0.664 \\
 & Non-Trainable & \textbf{0.870} & \textbf{0.761} \\
\hline
\multirow{6}{*}{BERT} & IndicBERT & 0.865 & 0.711 \\
 & mBERT & 0.903 & 0.783 \\
 & xlm-RoBERTa & 0.894 & 0.787 \\
 & MahaALBERT & 0.883 & 0.764 \\
 & MahaBERT & \textbf{0.909} & \textbf{0.803} \\
 & MahaRoBERTa & 0.902 & 0.803 \\
\hline
\end{tabular}
\caption{\label{table:2}
Classification results on different architectures}
\end{table*}

\begin{figure*}[hbt!]
\centering
\begin{subfigure}[b]{0.39\textwidth}
\includegraphics[width=\textwidth]{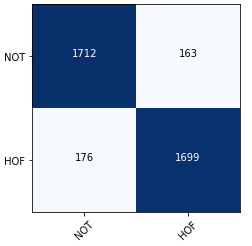}
\caption{2-class classification}
\label{fig:conf2}
\end{subfigure}
\begin{subfigure}[b]{0.4\textwidth}
\includegraphics[width=\textwidth]{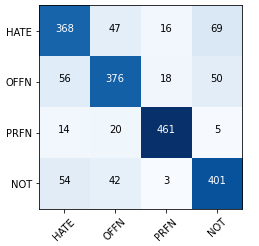}
\caption{4-class classification}
\label{fig:conf4}
\end{subfigure}
\caption{Confusion matrices for the best models}
\end{figure*}

\subsection{Results}
We performed our experiments on CNN, LSTM, and Transformer based models. For CNN and LSTM models, we have used random and fast text initialization for the word embeddings. The pre-trained embeddings were used in both trainable and non-trainable modes. The former means it was used by letting the embedding layer adapt to the training data and latter by preventing it from being updated during training. Additionally, we used pre-trained language models, particularly the variations of BERT such as IndicBERT, Multilingual BERT, XLM-RoBERTa, and a few custom BERT models to obtain the results. All the 2-class and 4-class accuracies are displayed in Table \ref{table:2}.
\\ \\
In CNN and LSTM based models, non-trainable fast text mode is outperforming other configurations in both the binary and 4-class results. All the monolingual Marathi BERT models are surpassing the multilingual versions of BERT models i.e IndicBERT, mBERT, and xlm-RoBERTa. It was observed that the non-trainable fast text setting for CNN and LSTM based models is performing competitively with the BERT models even surpassing the indicBERT for both the classes. The MahaBERT model gives the best binary classification results whereas MahaRoBERTa gives the best 4-class accuracy. The confusion matrices for respective best results  are shown in figures \ref{fig:conf2} and \ref{fig:conf4}.
% \begin{figure}[htp]
%     \centering
%     \includegraphics[width=.4\textwidth]{Images/conf2.png}\hfill
%     \includegraphics[width=.4\textwidth]{Images/conf4.png}\hfill
%     \caption{Distribution of the length of a tweet}
%     \label{fig:length}
% \end{figure}

\begin{figure*}[hbt!]
    \centering
    \includegraphics[width=\textwidth]{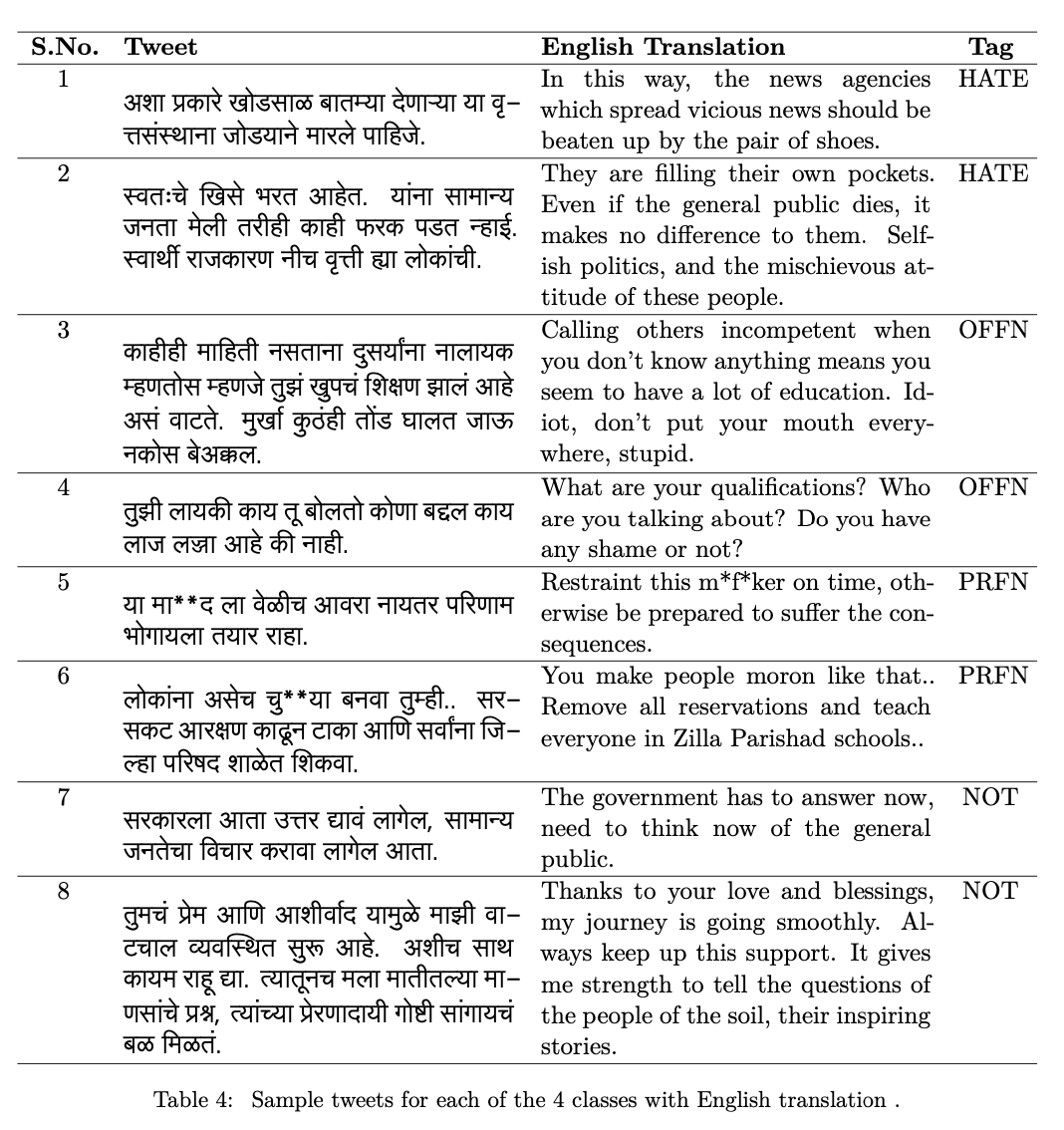}
    % \caption{Average characters and words per label}
    \label{fig:samples}
\end{figure*}
\section{Conclusion}

In this paper, we have presented L3CubeMahaHate - a hate speech dataset containing ~25000 distinct samples equally distributed in 4 classes. This is the first major dataset in the domain of hate speech. We also provide the binary version of the dataset of over ~37500 samples.
%We also explained the data collection process and annotation policies that we used throughout the curation of the dataset.
We further perform experiments to obtain baseline results on various deep learning models like CNN, LSTM, BiLSTM, and transformer-based BERT models such as IndicBERT, mBERT and RoBERTa. The dataset is also evaluated on monolingual Marathi BERT models like MahaBERT, MahaALBERT, and MahaRoBERTa. For CNN and LSTM based models, the non-trainable fast text mode outperforms its trainable counterpart in both binary and 4-class classification. In transformer-based models, MahaBERT and MahaRoBERTa give the best results in binary and 4-class classification respectively.

\section*{Acknowledgements} This work was done under the L3Cube Pune mentorship program. We would like to express our gratitude towards our mentors at L3Cube for their continuous support and encouragement.

\bibliographystyle{splncs04}
\bibliography{main}
\end{document}